\documentclass[sigconf]{acmart}

\usepackage{balance}

\def\BibTeX{{\rm B\kern-.05em{\sc i\kern-.025em b}\kern-.08emT\kern-.1667em\lower.7ex\hbox{E}\kern-.125emX}}

\copyrightyear{2019} 
\acmYear{2019} 
\setcopyright{acmcopyright}
\acmConference[KDD '19]{The 25th ACM SIGKDD Conference on Knowledge Discovery and Data Mining}{August 4--8, 2019}{Anchorage, AK, USA}
\acmBooktitle{The 25th ACM SIGKDD Conference on Knowledge Discovery and Data Mining (KDD '19), August 4--8, 2019, Anchorage, AK, USA}
\acmPrice{15.00}
\acmDOI{10.1145/3292500.3330948}
\acmISBN{978-1-4503-6201-6/19/08}
    
\usepackage{amsmath}
\usepackage{amsfonts}
\usepackage{algorithm}
\usepackage{algpseudocode}
\usepackage{color}
\usepackage{graphicx}
\usepackage{subfigure}
\usepackage{lipsum}
\usepackage{bm}
\usepackage{bbm}
\usepackage{tabu}
\usepackage{subeqnarray}
\usepackage{algorithm}
\usepackage{amsthm}
\usepackage{caption}
\usepackage{mathrsfs}
\usepackage{url}
\usepackage{booktabs}
\usepackage{diagbox}
\usepackage{pgfplots}
\usepackage{soul}
\usepackage{enumitem}
\begin{document}
\fancyhead{}
\bibliographystyle{plain}
%
\title{A Representation Learning Framework for Property Graphs}

\author{Yifan Hou, Hongzhi Chen, Changji Li, James Cheng, Ming-Chang Yang}
\affiliation{
	\institution{Department of Computer Science and Engineering}
	\institution{The Chinese University of Hong Kong}}
\email{{yfhou, hzchen, cjli, jcheng, mcyang}@cse.cuhk.edu.hk}

\renewcommand{\shortauthors}{Yifan Hou et al.}

%


%
\begin{abstract}

Representation learning on graphs, also called graph embedding, has demonstrated its significant impact on a series of machine learning applications such as classification, prediction and recommendation. However, existing work has largely ignored the rich information contained in the properties (or attributes) of both nodes and edges of graphs in modern applications, e.g., those represented by property graphs. To date, most existing graph embedding methods either focus on plain graphs with only the graph topology, or consider properties on nodes only. We propose PGE, a graph representation learning framework that incorporates both node and edge properties into the graph embedding procedure. PGE uses node clustering to assign biases to differentiate neighbors of a node and leverages multiple data-driven matrices to aggregate the property information of neighbors sampled based on a biased strategy. PGE adopts the popular inductive model for neighborhood aggregation. We provide detailed analyses on the efficacy of our method and validate the performance of PGE by showing how PGE achieves better embedding results than the state-of-the-art graph embedding methods on benchmark applications such as node classification and link prediction over real-world datasets.

\end{abstract}
%

%
\keywords{graph neural networks, graph embedding, property graphs, representation learning}

%
\maketitle

\section{Introduction}  \label{sec:intro}


Graphs are ubiquitous today due to the flexibility of using graphs to model data in a wide spectrum of applications. In recent years, more and more machine learning applications conduct classification or prediction based on graph data~\cite{euclidean_data, relational_machine_learning, overview_Standford,graph_embedding}, such as classifying protein's functions in biological graphs, understanding the relationship between users in online social networks, and predicting purchase patterns in buyers-products-sellers graphs in online e-commerce platforms. However, it is not easy to directly make use of the structural information of graphs in these applications as graph data are high-dimensional and non-Euclidean. On the other hand, considering only graph statistics such as degrees~\cite{introduction_degree}, kernel functions~\cite{introduction_kernel}, or local neighborhood structures~\cite{introduction_local_neighborhood} often provides limited information and hence affects the accuracy of classification/prediction.


Representation learning methods~\cite{representation_learning} attempt to solve the above-mentioned problem by constructing an embedding for each node in a graph, i.e., a mapping from a node to a low-dimensional Euclidean space as vectors, which uses geometric metrics (e.g., Euclidean distance) in the embedding space to represent the structural information. Such graph embeddings~\cite{graph_embedding, overview_Standford} have achieved good performance for classification/prediction on \textit{plain graphs} (i.e., graphs with only the pure topology, without node/edge labels and properties). However, in practice, most graphs in real-world do not only contain the topology information, but also contain labels and \emph{properties} (also called \emph{attributes}) on the entities (i.e., nodes) and relationships (i.e., edges). For example, in the companies that we collaborate with, most of their graphs (e.g., various graphs related to products, buyers and sellers from an online e-commerce platform;  mobile phone call networks and other communication networks from a service provider) contain rich node properties (e.g., user profile, product details) and edge properties (e.g., transaction records, phone call details). We call such graphs as \textbf{property graphs}. Existing methods~\cite{DeepWalk,node2vec,Walklets,HARP,SDNE,features_nn_3,graphsage} have not considered to take the rich information carried by both nodes and edges into the graph embedding procedure.


This paper studies the problem of property graph embedding. There are two main challenges. First, each node $v$ may have many properties and it is hard to find which properties may have greater influence on $v$ for a specific application. For example, consider the classification of papers into different topics for a citation graph where nodes represent papers and edges model citation relationships. Suppose that each node has two properties, ``year'' and ``title''. Apparently, the property ``title'' is likely to be more important for paper classification than the property ``year''. Thus, how to measure the influence of the properties on each node for different applications needs to be considered. Second, for each node $v$, its neighbors, as well as the connecting edges, may have different properties. How to measure the influences of both the neighbors and the connecting edges on $v$ for different application poses another challenge. In the above example, for papers referencing a target paper, those with high citations should mean more to the target paper than those with low citations.

Among existing work, GCN~\cite{features_nn_3} leverages node property information for node embedding generation, while GraphSAGE~\cite{graphsage} extends GCN from a spectral method to a spatial one. Given an application, GraphSAGE trains a weight matrix before embedding and then aggregates the property information of the neighbors of each node with the trained matrix to compute the node embedding. However, GraphSAGE does not differentiate neighbors with property dissimilarities for each node, but rather treats all neighbors equally when aggregating their property information. Moreover, GraphSAGE considers only node information and ignores edge directions and properties. Apart from the properties on nodes/edges, real-world graphs also have special structural features. For example, in social networks, nodes are often organized in the form of communities, where similar nodes are either neighbors due to the~\textit{homophily} feature~\cite{homophily_reason_1,homophily_reason_2}, or not direct neighbors but with similar structure due to the \textit{structural equivalent} feature~\cite{structural_equivalent_reason_1,structural_equivalent_reason_2,structural_equivalent_reason_3}. Thus, it is important to also consider structural features. For that, node2vec~\cite{node2vec} learns node embeddings by combining two strategies, breadth-first random walk and depth-first random walk, to account for the homophily feature and structural equivalent feature. However, node2vec only utilizes these two structural features without considering any property information.

To address the limitations of existing methods, we propose a new framework, \textbf{PGE}, for property graph embedding. PGE applies a biased method to differentiate the influences of the neighbors and the corresponding connecting edges by incorporating both the topology and property information into  the graph embedding procedure. The framework consists of three main steps: (1)~\textit{property-based node clustering} to classify the neighborhood of a node into similar and dissimilar groups based on their property similarity with the node; (2)~\textit{biased neighborhood sampling} to obtain a smaller neighborhood  sampled according the bias parameters (which are set based on the clustering result), so that the embedding process can be more scalable; and (3)~\textit{neighborhood aggregation} to compute the final low dimensional node embeddings by aggregating the property information of sampled neighborhood with weight matrices trained with neural networks. We also analyze in details how the three steps work together to contribute to a good graph embedding and why our biased method (incorporating node and edge information) can achieve better embedding results than existing methods.

We validated the performance of PGE by comparing with representative graph embedding methods, including DeepWalk~\cite{DeepWalk} and node2vec~\cite{node2vec} representing \textit{random walk based methods}, GCN~\cite{features_nn_3} for \textit{graph convolutional networks}, and GraphSAGE~\cite{graphsage} for \textit{neighbor aggregation based on weight matrices}. We tested these methods for two benchmark applications, node classification and link prediction, on a variety of real-world graphs. The results show that PGE achieves significant performance improvements over these existing methods. The experimental evaluation validates the importance of incorporating node/edge property information, in addition to topology information, into graph embedding. It also demonstrates the effectiveness of our biased strategy that differentiates neighbors to obtain better embedding results.


\section{Related Work}	\label{sec:related}
There are three main methods for graph embedding: \textit{matrix factorization}, \textit{random walk}, and \textit{neighbors aggregation}.


For matrix factorization methods, \cite{factorization_GF, factorization_GraRep} use adjacency matrix to define and measure the similarity among nodes for graph embedding. HOPE~\cite{factorization_HOPE} further preserves high-order proximities and obtains asymmetric transitivity for directed graphs. Another line of works utilize the random walk statistics to learn embeddings with the skip-gram model~\cite{skip_gram}, which applies vector representation to capture word relationships. 

The key idea of random walk is that nodes usually tend to co-occur on short random walks if they have similar embeddings~\cite{overview_Standford}. DeepWalk~\cite{DeepWalk} is the first to input random walk paths into a skip-gram model for learning node embeddings. node2vec~\cite{node2vec} further utilizes biased random walks to improve the mapping of nodes to a low-dimensional space, while combining breadth-first walks and depth-first walks to consider graph homophily and structural equivalence. To obtain larger relationships, Walklets~\cite{Walklets} involves \textit{offset} to allow longer step length during a random walk, while HARP~\cite{HARP} makes use of graph preprocessing that compresses some nodes into one super-node to improve random walk. 

According to~\cite{overview_Standford}, \textit{matrix factorization} and \textit{random walk} methods are \textit{shallow embedding approaches} and have the following drawbacks. First, since the node embeddings are independent and there is no sharing of parameters or functions, these methods are not efficient for processing large graphs. Second, they do not consider node/edge properties. Third, as the embeddings are transductive and can only be generated during the training phrase, unseen nodes cannot be embedded with the model being learnt so far.

To address (some of) the above problems, graph-based neural networks have been used to learn node embeddings, which encode nodes into vectors by compressing neighborhood information~\cite{deep_neural_network, DNGR, SDNE}. However, although this type of methods can share parameters, strictly speaking they are still transductive and have performance bottlenecks for processing large graphs as the input dimensionality of auto-encoders is equal to the number of nodes. Several recent works~\cite{graphsage, features_nn_1, features_nn_2, features_nn_3, NB-GCN} attempted to use only local neighborhood instead of the entire graph to learn node embeddings through neighbor aggregation, which can also consider property information on nodes. GCN~\cite{features_nn_3} uses graph convolutional networks to learn node embeddings, by merging local graph structures and features of nodes to obtain embeddings from the hidden layers. GraphSAGE~\cite{graphsage} is inductive and able to capture embeddings for unseen nodes through its trained auto-encoders directly. The advantage of neighborhood aggregation methods is that they not only consider the topology information, but also compute embeddings by aggregating property vectors of neighbors. However, existing neighborhood aggregation methods treat the property information of neighbors equally and fail to differentiate the influences of neighbors (and their connecting edges) that have different properties.
\section{The PGE Framework}	\label{sec:framework}


We use $\mathcal{G} = \{\mathcal{V},\mathcal{E}, \mathcal{P}, \mathcal{L}\}$ to denote a property graph , where $\mathcal{V}$ is the set of nodes and $\mathcal{E}$ is the set of edges.  $\mathcal{P}$ is the set of all properties and   $\mathcal{P} = \mathcal{P}_\mathcal{V} \cup \mathcal{P}_\mathcal{E}$, where $\mathcal{P}_\mathcal{V} = \bigcup_{v \in \mathcal{V}} \{ p_v \}$, \ $\mathcal{P}_\mathcal{E} = \bigcup_{e \in \mathcal{E}} \{ p_e \}$, and $p_v$ and $p_e$ are the set of properties of node $v$ and edge $e$, respectively.  $\mathcal{L} = \mathcal{L}_\mathcal{V} \cup \mathcal{L}_\mathcal{E}$ is the set of labels, where $\mathcal{L}_\mathcal{V}$ and $\mathcal{L}_\mathcal{E}$ are the sets of node and edge labels, respectively.  We use $\mathcal{N}_v$ to denote the set of neighbors of node $v \in \mathcal{V}$, i.e., $\mathcal{N}_v = \{v': (v,v') \in \mathcal{E}\}$. In the case that $\mathcal{G}$ is directed, we may further define $\mathcal{N}_v$ as the set of in-neighbors and the set of out-neighbors, though in this paper we abuse the notation a bit and do not use new notations such as $\mathcal{N}^{in}_v$ and $\mathcal{N}^{out}_v$ for simplicity of presentation, as the meaning should be clear from the context.

The property graph model is general and can represent other popular graph models. If we set $\mathcal{P} = \emptyset$ and $\mathcal{L} = \emptyset$, then $\mathcal{G}$ becomes a \textit{plain graph}, i.e., a graph with only the topology.  If we set $\mathcal{P}_\mathcal{V} = \mathcal{A}$, $\mathcal{P}_\mathcal{E} = \emptyset$, and $\mathcal{L} = \emptyset$, where $\mathcal{A}$ is the set of node attributes, then $\mathcal{G}$ becomes an \textit{attributed graph}. If we set $\mathcal{L} = \mathcal{L}_{\mathcal{V}}$, $\mathcal{P} = \emptyset$, and $\mathcal{L}_{\mathcal{E}} = \emptyset$, then $\mathcal{G}$ is a \textit{labeled graph}. 

\subsection{Problem Definition}	\label{framework:def}

The main focus of PGE is to utilize both topology and property information in the embedding learning procedure to improve the results for different applications. Given a property graph $\mathcal{G} = \{\mathcal{V},\mathcal{E},\mathcal{P},\mathcal{L}\}$, we define the similarity between two nodes $v_i, v_j \in \mathcal{V}$ as $s_{\mathcal{G}}(v_i,v_j)$.
The similarity can be further decomposed into two parts, $s_{\mathcal{G}}(v_i,v_j) = l(s_{\mathcal{P}}(v_i,v_j), s_{\mathcal{T}}(v_i,v_j))$, where $s_{\mathcal{P}}(v_i,v_j)$ is the property similarity and $s_{\mathcal{T}}(v_i,v_j)$ is the topology similarity between $v_i$ and $v_j$, and $l(\cdot,\cdot)$ is a non-negative mapping.

The \textit{embedding} of node $v \in \mathcal{V}$ is denoted as $\textbf{z}_v$, which is a vector obtained by an \textit{encoder} ${\rm ENC}(v) = \textbf{z}_v$. Our objective is to find the optimal ${\rm ENC}(\cdot)$, which minimizes the gap $\sum_{v_i, \! v_j \in \mathcal{V}} \! ||s_{\mathcal{G}}(v_i,v_j) - \textbf{z}_{v_i}^{\top} \textbf{z}_{v_j}|| \!=\! \sum_{v_i, \! v_j \in \mathcal{V}} \! ||  l(s_{\mathcal{P}}(v_i,v_j),  s_{\mathcal{T}}(v_i,v_j))  - \textbf{z}_{v_i}^{\top} \textbf{z}_{v_j}||$.

From the above problem definition, it is apparent that only considering the topology similarity $s_{\mathcal{T}}(v_i,v_j)$, as the traditional approaches do, cannot converge to globally optimal results. In addition, given a node $v$ and its neighbors $v_i, v_j$,  the property similarity $s_{\mathcal{P}}(v,v_i)$ can be very different from  $s_{\mathcal{P}}(v,v_j)$. Thus, in the PGE framework, we use both  topology similarity and  property similarity in learning the node embeddings.

\subsection{The Three Steps of PGE}\label{framework:steps}

The PGE framework  consists of three major steps as follows.

\begin{itemize}[leftmargin=*]
	\item Step 1: \textbf{Property-based Node Clustering.} We cluster nodes in $\mathcal{G}$  based on their properties to produce $k$ clusters $\mathcal{C}$$=$$\{ C_1, C_2,..., C_k \}$. A standard clustering algorithm such as $K$-Means~\cite{k_means} or DBSCAN~\cite{DBSCAN} can be used for this purpose, where each node to be clustered is represented by its property vector (note that graph topology information is not considered in this step).


	\item Step 2: \textbf{Biased Neighborhood Sampling.} To combine the influences of property information and graph topology by $l(\cdot,\cdot)$, we conduct biased neighborhood sampling based on the results of clustering in Step 1. To be specific, there are two phases in this step: (1)~For each neighbor $v'\in \mathcal{N}_v$, if $v'$ and $v$ are in the same cluster, we assign a \textit{bias} $b_s$ to $v'$ to indicate that they are similar; otherwise we assign  a different \textit{bias} $b_d$ to $v'$ instead to indicate that they are dissimilar. \ \ (2)~We normalize the assigned biases on $\mathcal{N}_v$, and then sample $\mathcal{N}_v$ according to the normalized biases to obtain a fixed-size sampled neighborhood $\mathcal{N}_v^s$.
	
	
	\item Step 3: \textbf{Neighborhood Aggregation.} Based on the sampled neighbors $\mathcal{N}_v^s$ in Step 2, we aggregate their property information to obtain $\textbf{z}_v$ by multiplying the weight matrices that are trained with neural networks.
\end{itemize}

In the following three sub-sections, we discuss the purposes and details of each of the above three steps.

\subsubsection{Property-based Node Clustering} 

The purpose of Step 1 is to classify $\mathcal{N}_v$ into two types for each node $v$ based on their node property information, i.e., those similar to $v$ or dissimilar to $v$. If $v$ and  its neighbor $v' \in \mathcal{N}_v$ are in the same cluster, we will regard $v'$ as a similar neighbor of $v$. Otherwise, $v'$ is dissimilar to $v$.

Due to the high dimensionality and sparsity of properties (e.g., property values are often texts but can also be numbers and other types), which also vary significantly from datasets to datasets, it is not easy to classify the neighborhood of each node into similar and dissimilar groups while maintaining a unified global standard for classifying the neighborhood of all nodes. For example, one might attempt to calculate the property similarity between $v$ and each of $v$'s neighbors, for all $v \in \mathcal{V}$, and then set a threshold to classify the neighbors into similar and dissimilar groups. However,  different nodes may require a different threshold and their similarity ranges can be very different. Moreover, each node's neighborhood may be classified differently and as we will show later, the PGE framework actually uses the 2-hop neighborhood while this example only considers the 1-hop neighborhood. Thus, we need a unified global standard for the classification. For this purpose, clustering the nodes based on their properties allows all nodes to be classified based on the same global standard. For example, the 1-hop neighbors and the 2-hop neighbors of a node $v$ are classified in the same way based on whether they are in the same cluster as $v$.

\subsubsection{Biased Neighborhood Sampling}

Many real-world graphs have high-degree nodes, i.e., these nodes have a large number of neighbors. It is inefficient and often unnecessary to consider all the neighbors for neighborhood aggregation in Step 3. Therefore, we use the biases $b_s$ and $b_d$ to derive a sampled neighbor set $\mathcal{N}_v^s$ with a fixed size for each node $v$. As a result, we obtain a sampled graph $\mathcal{G}^s = \{ \mathcal{V}, \mathcal{E}^s \}$, where $\mathcal{E}^s = \{ (v,v') : v' \in \mathcal{N}_v^s \}$. Since  the biases $b_s$ and $b_d$ are assigned to the neighbors based on the clustering results computed from the node properties, $\mathcal{G}^s$ contains the topology information of $\mathcal{G}$ while it is constructed based on the node property information. Thus, Step 2 is essentially a mapping $l(\cdot,\cdot)$  that fuses  $s_{\mathcal{P}}(v, v')$ and $s_{\mathcal{T}}(v, v')$. 

The biases $b_s$ and $b_d$ are the un-normalized possibility of selecting neighbors from dissimilar and similar clusters, respectively. The value of $b_s$ is set to 1, while $b_d$ can be varied depending on the probability (greater $b_d$ means higher probability) that dissimilar neighbors should be selected into $\mathcal{G}^s$. We will analyze the effects of the bias values in Section~\ref{sec:analysis} and verify by experimental results in Section~\ref{exp:bias}. The size of $\mathcal{N}_v^s$ is set to $25$ by default following GraphSAGE~\cite{graphsage} (also for fair comparison in our experiments). The size $25$ was found to be a good balance point in~\cite{graphsage} as a larger size will significantly increase the model computation time, though in the case of PGE as it differentiates neighbors, using a sampled neighborhood could achieve a better quality of embedding than using the full neighborhood.



\subsubsection{Neighborhood Aggregation}	\label{framework:aggregation}

The last step is to learn the low dimensional embedding with $\mathcal{G}^s = \{ \mathcal{V}, \mathcal{E}^s \}$. We use neighborhood aggregation to learn the function ${\rm ENC}(\cdot)$ for generating the node embeddings. For each node, we select its neighbors within two hops to obtain $\textbf{z}_{v}$ by the following equations:
$$ \textbf{z}_{v} = \sigma(W^1 \cdot {\rm {A}} (\textbf{z}_{v}^1, \sum_{v' \in \mathcal{N}_v^s}{\textbf{z}_{v'}^1}/{|\mathcal{N}_v^s|} )),$$
$$\textbf{z}_{v'}^1 = \sigma(W^2 \cdot {\rm {A}} (p_{v'}, \sum_{v'' \in \mathcal{N}_{v'}^s}{p_{v''}}/{|\mathcal{N}_{v'}^s|} )),$$
\noindent where $p_v$ is the original property vector of node $v$, \textbf{$\sigma(\cdot)$} is the nonlinear activation function and \textbf{${\rm {A}}(\cdot)$} is the \textit{concatenate} operation. We use two weight matrices $W^1$ and $W^2$ to aggregate the node property information of $v$'s one-hop neighbors and two-hop neighbors.

The matrix $W^i$ is used to assign different weights to different properties because aggregating (e.g., taking mean value) node property vectors directly cannot capture the differences between properties, but different properties contribute to the embedding in varying degrees. Also, the weight matrix is data-driven and should be trained separately for different datasets and applications, since nodes in different graphs have different kinds of properties. The weight matrices are pre-trained using Adam SGD optimizer~\cite{adam}, with a loss function defined for the specific application, e.g., for node classification, we use binary cross entropy loss (multi-labeled); for link prediction, we use cross entropy loss with negative sampling.

\subsection{Support of Edge Direction and Properties}	\label{framework:edge}


The sampled graph $\mathcal{G}^s$ does not yet consider the edge direction and edge properties. To include edge properties, we follow the same strategy as we do on nodes. If edges are directed, we consider in-edges and out-edges separately. We cluster the edges into $k^e$ clusters $\mathcal{C}^e = \{ C_1^e, C_2^e,..., C_{k^e}^e \}$. Then, we train $2 \cdot k^e$ matrices, $\{W^1_{1},W^1_{2},...,W^1_{k^e}\}$ and $\{W^2_{1},W^2_{2},...,W^2_{k^e}\}$, to aggregate node properties for $k^e$ types of edges for the 2-hop neighbors. Finally, we obtain $\textbf{z}_v$ by the following equations:
$$ \textbf{z}_{v} \!=\! \sigma\Big( {\rm {A}} \big(W^1_0 \cdot \textbf{z}_{v}^1, {\rm {A}}_{C_{i}^e \in \mathcal{C}^e} (W^1_i \cdot
\mathbb{E}_{v' \in \mathcal{N}_v^s \& \! \ (v,v') \in C_i^e }
{ [\textbf{z}_{v'}^1]}) \big)\Big) , \eqno{(1)}$$
$$ \textbf{z}_{v'}^{1} \!=\! \sigma\Big( {\rm {A}} \big(W^2_0 \! \cdot \! p_{v'}, {\rm {A}}_{C_{i}^e \in \mathcal{C}^e} (W^2_i \! \cdot \!
\mathbb{E}_{v'' \in \mathcal{N}_{v'}^s \& \! \ (v',v'') \in C_i^e }
{ [p_{v''}]}) \big)\Big) . \eqno{(2)} $$
Note that $|\mathcal{C}^e|$ should not be too large as to avoid high-dimensional vector operations. Also, if $|\mathcal{C}^e|$ is too large, some clusters may contain only a few elements, leading to under-fitting for the trained weight matrices. Thus, we set $|\mathcal{C}^e|$ as a fixed small number.




\subsection{The Algorithm}

{\small
\renewcommand{\algorithmicrequire}{\textbf{Input:}}  
\renewcommand{\algorithmicensure}{\textbf{Output:}} 
\begin{algorithm}[!t]
	\caption{Property Graph Embedding (PGE)}
	\label{alg:framwork}
	\begin{algorithmic}[1]	
		\Require
		A Property Graph $\mathcal{G}=\{ \mathcal{V}, \mathcal{E}, \mathcal{P} \}$;
		biases $b_d$ and $b_s$; the size of sampled neighborhood $|\mathcal{N}_v^s|$;
		weight matrices $\{W^1_{1},W^1_{2},...,W^1_{k^e}\}$ and $\{W^2_{1},W^2_{2},...,W^2_{k^e}\}$
		\Ensure
		Low-dimensional representation vectors $\textbf{z}_v$, $\forall v \in \mathcal{V}$
		\State
		Clustering $\mathcal{V}$, $\mathcal{E}$, and obtain $\mathcal{C}$ and $\mathcal{C}^e$ based on $\mathcal{P}$;
		\Comment{step 1}{}
		
		\ForAll {$v \in \mathcal{V}$}
		\Comment{step 2}{}
		\ForAll {$v' \in \mathcal{N}_v$}
		\State
		Assign $b=b_d+(b_s-b_d) \cdot \sum_{C_i \in \mathcal{C}} \mathbb{I}\{ v,v' \in C_i \} $ to $v'$, 
		\State
		where $\mathbb{I}\{ v,v' \in C_i \}=1$ if $v,v' \in C_i$ and 0 otherwise;
		\EndFor
		\State
		Sample $|\mathcal{N}_v^s|$ neighbors with bias $b$;
		\EndFor
		
		\ForAll {$v \in \mathcal{V}$}
		\Comment{step 3}{}
		\State
		Compute $\textbf{z}_v^1$ with Equation $(2)$;
		\EndFor
		\ForAll {$v \in \mathcal{V}$}
		\State
		Compute $\textbf{z}_v$ with Equation $(1)$;
		\EndFor
	\end{algorithmic}
\end{algorithm}
}


Algorithm~\ref{alg:framwork} presents the overall procedure of computing the embedding vector $\textbf{z}_v$ of each node $v \in \mathcal{V}$. The algorithm follows exactly the three steps that we have described in Section~\ref{framework:steps}. 

\section{An Analysis of PGE}	\label{sec:analysis}


In this section, we present a detailed analysis of PGE. In particular, we analyze why the biased strategy used in PGE can improve the embedding results. We also discuss how the bias values $b_d$ and $b_s$ and edge information affect the embedding performance. 


\subsection{The Efficacy of the Biased Strategy} \label{analysis:strategy}

One of the main differences between PGE and GraphSAGE~\cite{graphsage} is that neighborhood sampling in PGE is biased (i.e., neighbors are selected based on probability values defined based on $b_d$ and $b_s$), while GraphSAGE's neighborhood sampling is unbiased (i.e., neighbors are sampled with equal probability). We analyze the difference between the biased and the unbiased strategies in the subsequent discussion.

We first argue that neighborhood sampling is a special case of random walk. For example, if we set the walk length to $1$ and perform $10$ times of walk, the strategy can be regarded as 1-hop neighborhood sampling with a fixed size of $10$. Considering that the random walk process in each step follows an i.i.d. process for all nodes, we define the biased strategy as a $|\mathcal{V}| \times |\mathcal{V}|$ matrix $\textbf{P}$, where $\textbf{P}_{i,j}$ is the probability that node $v_i$ selects its neighbor $v_j$ in the random walk. If two nodes $v_i$ and $v_j$ are not connected, then $\textbf{P}_{i,j}=0$. Similarly, we define the unbiased strategy $\textbf{Q}$, where all neighbors of any node have the same probability to be selected. We also assume that there exists an optimal strategy $\textbf{B}$, which gives the best embedding result for a given application. 


A number of works~\cite{node2vec, Walklets, HARP} have already shown that adding preference on similar and dissimilar neighbors during random walk can improve the embedding results, based on which we have the following statement: \textit{for a biased strategy $\textbf{P}$, if  $||\textbf{B}-\textbf{P}||_1 < ||\textbf{B}-\textbf{Q}||_1$, where $\textbf{B} \neq \textbf{Q}$, then  $\textbf{P}$ has a positive influence on improving the embedding results.}


Thus, to verify the efficacy of PGE's biased strategy, we need to show that our strategy $\textbf{P}$ satisfies  $ ||\textbf{B}-\textbf{P}||_{1} \leq ||\textbf{B}-\textbf{Q}||_{1}$. To do so, we show that $b_d$ and $b_s$ can be used to adjust the strategy $\textbf{P}$ to get closer to $\textbf{B}$ (than $\textbf{Q}$). 

Assume that nodes are classified into $k$ clusters $\mathcal{C}= \{C_1, C_2,...,C_{k}\}$ based on the property information $\mathcal{P}_{\mathcal{V}}$. For the unbiased strategy, the expected similarity of two nodes $v, v' \in \mathcal{V}$ for each random walk step is:
$$\mathbb{E}[s_{\mathcal{G}(v,v')}] = \frac {\sum_{v \in \mathcal{V}} \sum_{v_i \in \mathcal{N}_v} s_{\mathcal{G}}(v,v_i)} {|\mathcal{E}|}.$$ 
The expectation of two nodes' similarity for each walk step in our biased strategy is:
$$\mathbb{E}[s_{\mathcal{G}(v,v')}] =  {\frac {\sum_{v \in \mathcal{V}} \sum_{v_i \in \mathcal{N}_v \cap C_v} n_s(v) \cdot s_{\mathcal{G}}(v,v_i)} {\frac {|\mathcal{E}|} {k}}} $$
$$
+  {\frac {\sum_{v \in \mathcal{V}} \sum_{v_j \in \mathcal{N}_v \cap (C_v)^c } n_d(v) \cdot s_{\mathcal{G}}(v,v_j)} {\frac {|\mathcal{E}| \cdot (k- 1)} {k}}}, \eqno{(3)}$$ 
\noindent where $n_s(v)$ and  $n_d(v)$ are the normalized biases of $b_s$ and  $b_d$ for node $v$ respectively, $C_v$ is the cluster that contains $v$, and $(C_v)^c = \mathcal{C}\backslash \{C_v\}$. Since only connected nodes are to be selected in a random walk step, the normalized biases $n_s(v)$ and $n_d(v)$ can be derived by 
$$n_s(v)=\frac {b_s}{b_d \cdot \sum_{v' \in \mathcal{N}_v }\mathbb{I}\{v' \in C_v\}  +b_s \cdot \sum_{v' \in \mathcal{N}_v }\mathbb{I}\{v' \in (C_v)^c\} },$$
\noindent and $$n_d(v) = n_s(v) \times \frac {b_d} {b_s}.$$ 

Consider Equation~$(3)$, if we set $b_d= b_s$, which means $n_d(v)=n_s(v)$, then it degenerates to the unbiased random walk strategy. But if we set $b_d$ and $b_s$ differently, we can adjust the biased strategy to either (1)~select more dissimilar neighbors by assigning $b_d > b_s$ or (2)~select more similar neighbors by assigning $b_s > b_d$.

Assume that the clustering result is not trivial, i.e., we obtain at least more than 1 cluster, we can derive that 
$$\frac {\sum_{C_i \in \mathcal{C}} \sum_{v,v' \in C_i} s_{\mathcal{P}}(v,v')} {\frac 1 2 \sum_{C_i \in \mathcal{C}} |C_i| \cdot (|C_i|-1)} > \frac {\sum_{v, v' \in \mathcal{V}} s_{\mathcal{P}}(v,v')} {\frac 1 2 |V| \cdot (|V|-1)}.$$
Since $l(\cdot,\cdot)$ is a non-negative mapping with respect to $s_{\mathcal{P}}(v,v')$,  we have
$$\frac {\sum_{C_i \in \mathcal{C}} \sum_{v,v' \in C_i} s_{\mathcal{G}}(v,v')} {\frac 1 2 \sum_{C_i \in \mathcal{C}} |C_i| \cdot (|C_i|-1)} > \frac {\sum_{v, v' \in \mathcal{V}} s_{\mathcal{G}}(v,v')} {\frac 1 2 |V| \cdot (|V|-1)} \eqno{(4)}.$$
Equation~$(4)$ shows that the similarity $s_{\mathcal{G}}(v,v')$ is higher if $v$ and $v'$ are in the same cluster. Thus, based on Equations~$(3)$ and~$(4)$, we conclude that parameters $b_d$ and $b_s$ can be used to select similar and dissimilar neighbors. 

Next, we consider the optimal strategy $\textbf{B}$ for 1-hop neighbors, where $\textbf{B}_{i,j} = \mathbb{I}\{ v_j \in \mathcal{N}_{v_i} \} \cdot b_{v_i,v_j}^*$, and $b_{v_i,v_j}^*$ is the normalized optimal bias value for $\textbf{B}_{i,j}$. Similarly, the unbiased strategy is $\textbf{Q}_{i,j} = \mathbb{I}\{ v_j \in \mathcal{N}_{v_i} \} \cdot \frac {1} {|\mathcal{N}_{v_i}|}$. Thus, we have
$$||\textbf{B}-\textbf{Q}||_{1} = \sum_{v_i \in \mathcal{V}} \sum_{v_j \in \mathcal{V}} \Big| b_{v_i,v_j}^* - \frac {1} {|\mathcal{N}_{v_i}|} \Big|.$$ 


For our biased strategy, $\textbf{P}_{i,j} = \mathbb{I}\{ v_j \in \mathcal{N}_{v_i} \cap C_{v_i} \} \cdot n_s(v) + \mathbb{I}\{ v_j \in \mathcal{N}_{v_i} \cap (C_{v_i})^c \} \cdot n_d(v)$. There exist $b_s$ and $b_d$ that satisfy $\sum_{v_i \in \mathcal{V}} \sum_{v_j \in \mathcal{V}} \Big| b_{v_i,v_j}^* - \frac {1} {|\mathcal{N}_{v_i}|} \Big| \ge \sum_{v_i \in \mathcal{V}} \sum_{v_j \in \mathcal{V}} \Big| b_{v_i,v_j}^* - \mathbb{I}\{ v_j \in \mathcal{N}_{v_i} \cap C_{v_i} \} \cdot n_s(v) - \mathbb{I}\{ v_j \in \mathcal{N}_{v_i} \cap (C_{v_i})^c \} \cdot n_d(v) \Big|$, where strict inequality can be derived if $b_d \neq b_s$.
Thus, $||\textbf{B}-\textbf{P}||_{1} < ||\textbf{B}-\textbf{Q}||_{1}$ if we set proper values for $b_s$ and $b_d$ (we discuss the bias values in Section~\ref{analysis:bias}). Without loss of generality, the above analysis can be extended to the case of multi-hop neighbors. 



\subsection{The Effects of the Bias Values}	\label{analysis:bias}

Next we discuss how to set the proper values for the biases $b_s$ and $b_d$ for neighborhood sampling. We also analyze the impact of the number of clusters on the performance of PGE.

For neighborhood aggregation in Step 3 of PGE, an accurate embedding of a node $v$ should be obtained by covering the whole connected component that contains $v$, where all neighbors within $k$-hops ($k$ is the maximum reachable hop) should be aggregated. However, for a large graph, the execution time of neighborhood aggregation increases rapidly beyond 2 hops, especially for power-law graphs. For this reason, we trade accuracy by considering only the 2-hop neighbors. In order to decrease the accuracy degradation, we can enlarge the change that a neighbor can contribute to the embedding $\textbf{z}_v$ by selecting dissimilar neighbors within the 2-hops, which we elaborate as follows. 

Consider a node $v\in \mathcal{V}$ and its two neighbors $v_i,v_j \in \mathcal{N}_v$, and assume that $\mathcal{N}_{v_i} = \mathcal{N}_{v_j}$ but $|p_v-p_{v_i}| < |p_v-p_{v_j}|$. Thus, we have $s_{\mathcal{T}}(v,v_i) = s_{\mathcal{T}}(v,v_j)$ and $s_{\mathcal{P}}(v,v_i) > s_{\mathcal{P}}(v,v_j)$. Since $l(\cdot,\cdot)$ is a non-negative mapping, we also have $s_{\mathcal{G}}(v,v_i) > s_{\mathcal{G}}(v,v_j)$. Based on the definitions of $\textbf{z}_{v}$ and $\textbf{z}_{v'}^1$ given in Section~\ref{framework:aggregation}, by expanding $\textbf{z}_{v'}^1$ in  $\textbf{z}_{v}$, we obtain 
$$ \textbf{z}_{v} \!=\! \sigma\bigg(W^1 \cdot {\rm {A}} \Big(\textbf{z}_{v}^1, \!\!\! \sum_{v' \in \mathcal{N}_v^s} \!\!\! \sigma \big(W^2 \cdot {\rm {A}} (p_{v'}, \!\!\!\! \sum_{  v'' \in \mathcal{N}_{v'}^s} \!\!\! {p_{v''}}/{|\mathcal{N}_{v'}^s|} ) \big)/{|\mathcal{N}_v^s|} \Big) \bigg). \eqno{(5)}$$ 
Equation~$(5)$ aggregates the node property vector $p_v$ (which is represented within $\textbf{z}_{v}^1$) and the property vectors of $v$'s 2-hop neighbors to obtain the node embedding $\textbf{z}_v$. This procedure can be understood as transforming from $s_{\mathcal{P}}(v,v')$ to $s_{\mathcal{G}}(v,v')$. Thus, a smaller $s_{\mathcal{P}}(v,v')$ is likely to contribute a more significant change to $\textbf{z}_v$. With Equation~$(5)$, if $|p_v - p_{v_i}| < |p_v - p_{v_j}|$, we obtain $||\textbf{z}_v^1 - \textbf{z}_{v_i}^1||_{1} < ||\textbf{z}_v^1 - \textbf{z}_{v_j}^1||_{1}$. Then, for the embeddings, we have $||\textbf{z}_v - \textbf{z}_{v_i}||_{1} < ||\textbf{z}_v - \textbf{z}_{v_j}||_{1}$. Since $v$ and $v_i$, as well as $v$ and $v_j$, have mutual influence on each other, we conclude that for fixed-hop neighborhood aggregation, the neighbors with greater dissimilarity can contribute larger changes to the node embeddings. That is, for fixed-hop neighborhood aggregation, we should set $b_d > b_s$ for better embedding results, which is also validated in our experiments.

Apart from the values of $b_d$ and $b_s$, the number of clusters obtained in Step 1 of PGE may also affect the quality of the node embeddings. Consider a random graph $\mathcal{G} = \{ \mathcal{V}, \mathcal{E}, \mathcal{P} \}$ with average degree $|\mathcal{E}|/|\mathcal{V}|$. Assume that we obtain $k$ clusters from  $\mathcal{G}$ in Step 1, then the average number of neighbors in $\mathcal{N}_v$ that are in the same cluster with a node $v$ is $|\mathcal{N}_v|/k = (|\mathcal{E}|/|\mathcal{V}|)/k$. If $k$ is large, most neighbors will be in different clusters from the cluster of $v$. On the contrary, a small $k$ means that neighbors in $\mathcal{N}_v$ are more likely to be in the same cluster as $v$. Neither an extremely large $k$ or small $k$ gives a favorable condition for node embedding based on the biased strategy because we will have either all dissimilar neighbors or all similar neighbors, which essentially renders the neighbors in-differentiable. Therefore, to ensure the efficacy of the biased strategy, the value of $k$ should not fall into either of the two extreme ends. We found that a value of $k$ close to the average degree is a good choice based on our experimental results.

\subsection{Incorporating Edge Properties} \label{analysis:edge}


In addition to the biased values and the clustering number, the edge properties can also bring significant improvements on the embedding results. Many real-world graphs such as online social networks have edge properties like ``positive'' and ``negative''. Consider a social network $\mathcal{G} = \{ \mathcal{V}, \mathcal{E}, \mathcal{P} \}$ with two types of edges, $\mathcal{E} = \mathcal{E^+} \cup \mathcal{E^-}$. Suppose that there is a node $v \in \mathcal{V}$ having two neighbors $v_i, v_j \in \mathcal{N}_v$, and these two neighbors have exactly the same property information $p_{v_i} = p_{v_j}$ and topology information $\mathcal{N}_{v_i}=\mathcal{N}_{v_j}$, but are connected to $v$ with different types of edges, i.e., $(v,v_i) \in \mathcal{E}^+$ and $ (v,v_j) \in \mathcal{E}^-$. If we only use Equation~$(5)$, then we cannot differentiate the embedding results of $v_i$ and $v_j$ ($\textbf{z}_{v_i}$ and $\textbf{z}_{v_j}$). This is because the edges are treated equally and the edge property information is not incorporated into the embedding results. In order to incorporate the edge properties, we introduce an extra matrix for each property. For example, in our case two additional matrices are used for the edge properties ``positive'' and ``negative''; that is, referring to Section~\ref{framework:edge}, we have $k^e = 2$ in this case. In the case of directed graphs, we further consider the in/out-neighbors separately with different weight matrices as we have discussed in Section~\ref{framework:edge}.

\section{Experimental Evaluation} \label{sec:exp}

We evaluated the performance of PGE using two benchmark applications, \textit{node classification} and \textit{link prediction}, which were also used in the evaluation of many existing graph embedding methods~\cite{overview_Standford,relational_machine_learning,graph_embedding}. In addition, we also assessed the effects of various parameters on the performance of PGE. 


\vspace{1mm}

\noindent \textbf{Baseline Methods.} \ We compared PGE with the representative works of the following three methods: \textit{random walk based on skip-gram}, \textit{graph convolutional networks}, and \textit{neighbor aggregation based on weight matrices}. 
\begin{itemize}
	\item DeepWalk~\cite{DeepWalk}: This work introduces the skip-gram model to learn node embeddings by capturing the relationships between nodes based on random walk paths. DeepWalk achieved significant improvements over its former works, especially for multi-labeled classification applications, and was thus selected for comparison.
	
	\item node2vec~\cite{node2vec}: This method considers both graph homophily and structural equivalence. We compared PGE with node2vec as it is the representative work for graph embedding based on biased random walks. 
	
	\item GCN~\cite{features_nn_3}: This method is the seminal work that uses convolutional neural networks to learn node embedding.
	
	\item GraphSAGE~\cite{graphsage}: GraphSAGE is the state-of-the-art graph embedding method and uses node property information in neighbor aggregation. It significantly improves the performance compared with former methods by learning the mapping function rather than embedding directly. 
	
\end{itemize}

To ensure fair comparison, we used the optimal default parameters of the existing methods. For DeepWalk and node2vec, we used the same parameters to run the algorithms, with window size set to $10$, walk length set to $80$ and number of walks set to $10$. Other parameters were set to their default values. For GCN, GraphSAGE and PGE, the learning rate was set to $0.01$. For node classification, we set the epoch number to $100$ (for GCN the early stop strategy was used), while for link prediction we set it to $1$ (for PubMed we set it to $10$ as the graph has a small number of nodes). The other parameters of GCN were set to their optimal default values. PGE also used the same default parameters as those of GraphSAGE such as the number of sampled layers and the number of neighbors.

\vspace{1mm}

\noindent \textbf{Datasets.} \ We used four real-world datasets in our experiments, including a citation network, a biological protein-protein interaction network and two social networks. 
\begin{itemize}
	\item \textbf{PubMed}~\cite{pubmed} is a set of articles (i.e., nodes) related to diabetes from the PubMed database, and edges here represent the citation relationship. The node properties are TF/IDF-weighted word frequencies and node labels are the types of diabetes addressed in the articles.
	
	\item \textbf{PPI}~\cite{ppi} is composed of $24$ protein-protein interaction graphs, where each graph represents a human tissue. Nodes here are proteins and edges are their interactions. The node properties include positional gene sets, motif gene sets and immunological signatures. The node labels are gene ontology sets. We used the processed version of~\cite{graphsage}.
	
	\item \textbf{BlogCatalog}~\cite{blogcatalog} is a social network where users select categories for registration. Nodes are bloggers and edges are relationships between them (e.g., friends). Node properties contain user names, ids, blogs and blog categories. Node labels are user tags.
	
	\item \textbf{Reddit}~\cite{graphsage} is an online discussion forum. The graph was constructed from Reddit posts. Nodes here are posts and they are connected if the same users commented on them. Property information includes the post title, comments and scores. Node labels represent the community. We used the sparse version processed in~\cite{graphsage}.
\end{itemize}

\begin{table}[!t]
	\caption{Dataset statistics}
	\label{tab:datasets}
	\vspace{-2mm}
	\scalebox{0.81}{
		\begin{tabular}{c|c|c|c|c|c}
			\hline
			\toprule
			Dataset & $|\mathcal{V}|$ & $|\mathcal{E}|$ & avg. degree & feature dim. & \# of classes \\
			\midrule
			\hline
			PubMed & 19,717 & 44,338 & 2.25 & 500 & 3\\
			PPI & 56,944 & 818,716 & 14.38 & 50 & 121\\
			BlogCatalog  & 55,814 & 1,409,112 & 25.25 & 1,000 & 60 \\
			Reddit & 232,965 & 11,606,919 & 49.82 & 602 & 41\\
			\bottomrule
			\hline
		\end{tabular}
	}
\end{table}

Table \ref{tab:datasets} shows some statistics of the datasets. To evaluate the performance of node classification of the algorithms on each dataset, the labels attached to nodes are treated as classes, whose number is shown in the last column. Note that each node in PPI and BlogCatalog may have multiple labels, while that in PubMed and Reddit has only a single label. The average degree (i.e., $|\mathcal{E}|/|\mathcal{V}|$) shows that the citation dataset PubMed is a sparse graph, while the other graphs have higher average degree. For undirected graphs, each edge is stored as two directed edges.

\begin{table*}
	\caption{Performance of node classification}
	\label{tab:f1_score}
		\vspace{-2mm}
	\scalebox{.8}{
		\begin{tabular}{|c|c|c|c|c|}
			\hline
			\toprule
			\diagbox{Alg.}{F1-Micro (\%)}{Datasets} & PubMed & PPI & BlogCatalog & Reddit \\
			\midrule
			\hline
			DeepWalk & 78.85 & 60.66 & 38.69 & - \\
			node2vec & 78.53 & 61.98 & 37.79 & - \\
			GCN  & 84.61 & - & - & - \\
			GraphSAGE & 88.08 & 63.41 & 47.22 & 94.93 \\
			\textbf{PGE} & \textbf{88.36} & \textbf{84.31} & \textbf{51.31} & \textbf{95.62} \\
			\bottomrule
			\hline
	\end{tabular}}
	\scalebox{.8}{
		\begin{tabular}{|c|c|c|c|c|}
			\hline
			\toprule
			\diagbox{Alg.}{F1-Macro (\%)}{Datasets} & PubMed & PPI & BlogCatalog & Reddit \\
			\midrule
			\hline
			DeepWalk & 77.41 & 45.19 & 23.73 & - \\
			node2vec & 77.08 & 48.57 & 22.94 & - \\
			GCN  & 84.27 & - & - & - \\
			GraphSAGE & 87.87 & 51.85 & 30.65 & 92.30 \\
			\textbf{PGE} & \textbf{88.24} & \textbf{81.69} & \textbf{37.22} & \textbf{93.29} \\
			\bottomrule
			\hline
	\end{tabular}}
\end{table*}

\subsection{Node Classification}\label{exp:nodec}

We first report the results for node classification. All nodes in a graph were divided into three types: training set, validation set and test set for evaluation. We used 70\% for training, 10\% for validation and 20\% for test for  all datasets except for PPI, which is composed of 24 subgraphs and we followed GraphSAGE~\cite{graphsage} to use about 80\% of the nodes (i.e., those in 22 subgraphs) for training and nodes in the remaining 2 subgraphs for validation and test. For the biases, we used the default values, $b_s = 1$ and $b_d=1000$, for all the datasets. For the task of node classification, the embedding result (low-dimensional vectors) satisfies $\textbf{z}_v \in \mathbb{R}^{d_l}$, where $d_l$ is the number of classes as listed in Table~\ref{tab:datasets}. The index of the largest value in $\textbf{z}_v$ is the classification result for single-class datasets. In case of multiple classes, the rounding function was utilized for processing $\textbf{z}_v$ to obtain the classification results. We used F1-score~\cite{f1_score}, which is a popular metric for multi-label classification, to evaluate the performance of classification. 


Table~\ref{tab:f1_score} reports the results, where the left table presents the F1-Micro values and the right table presents the F1-Macro values. PGE achieves higher F1-Micro and F1-Macro scores than all the other methods for all datasets, especially for PPI and BlogCatalog for which the performance improvements are significant. In general, the methods that use node property information (i.e., PGE, GraphSAGE and GCN) achieve higher scores than the methods that use the skip-gram model to capture the structure relationships (i.e., DeepWalk and node2vec). This is because richer property information is used by the former methods than  the latter methods that use only the pure graph topology. Compared with GraphSAGE and GCN, PGE further improves the classification accuracy by introducing biases to differentiate neighbors for neighborhood aggregation, which validates our analysis on the importance of our biased strategy in Section~\ref{sec:analysis}. In the remainder of this subsection, we discuss in greater details the performance of the methods on each dataset.


To classify the article categories in PubMed, since the number of nodes in this graph is not large, we used the DBSCAN clustering method in Step 1, which produced $k=4$ clusters. Note that the graph has a low average degree of only $2.25$. Thus, differentiating the neighbors does not bring significant positive influence. Consequently, PGE's F1-scores are not significantly higher than those of GraphSAGE for this dataset. 


To classify proteins' functions of PPI, since this graph is not very large, we also used DBSCAN for clustering, which produced $k=39$ clusters. For this dataset, the improvement made by PGE over other methods is impressive, which could be explained by that neighbors in a protein-protein interaction graph play quite different roles and thus differentiating them may bring significantly benefits for node classification. In fact, although GraphSAGE also uses node property information, since GraphSAGE does not differentiate neighbors, it does not obtain significant improvement over DeepWalk and node2vec (which use structural information only). The small improvement made by GraphSAGE compared with the big improvement made by PGE demonstrates the effectiveness of our biased neighborhood sampling strategy. For GCN, since it does not consider multi-labeled classification, comparing it with the other methods is unfair and not meaningful for this dataset (also for BlogCatalog).

BlogCatalog has high feature dimensionality. The original BlogCatalog dataset regards the multi-hot vectors as the feature vectors (with $5,413$ dimensions). We used Truncate-SVD to obtain the low-dimensional feature vectors (with $1,000$ dimensions). Since the number of nodes is not large, we used DBSCAN for Step 1, which produced $k=18$ clusters for this dataset. The improvement in the classification accuracy made by PGE is very significant compared with DeepWalk and node2vec, showing the importance of using property information for graph embedding. The improvement over GraphSAGE is also quite significant for this dataset, which is due to both neighbor differentiation and the use of edge direction.

\begin{table}
	\caption{Performance of link prediction}
	\label{tab:mrr}
		\vspace{-2mm}
	\scalebox{0.8}{
		\begin{tabular}{c|c|c|c|c}
			\hline
			\toprule
			\diagbox{Alg.}{MRR (\%)}{Datasets} & PubMed & PPI & BlogCatalog & Reddit \\
			\midrule
			\hline
			GraphSAGE & 43.72 & 39.93 & 24.61 & 41.27 \\
			PGE (no edge info) & 41.47 & 59.73 & 23.89 & 39.81 \\
			\textbf{PGE} & \textbf{70.77} & \textbf{89.21} & \textbf{72.97} & \textbf{56.59} \\
			\bottomrule
			\hline
	\end{tabular}}
\end{table}

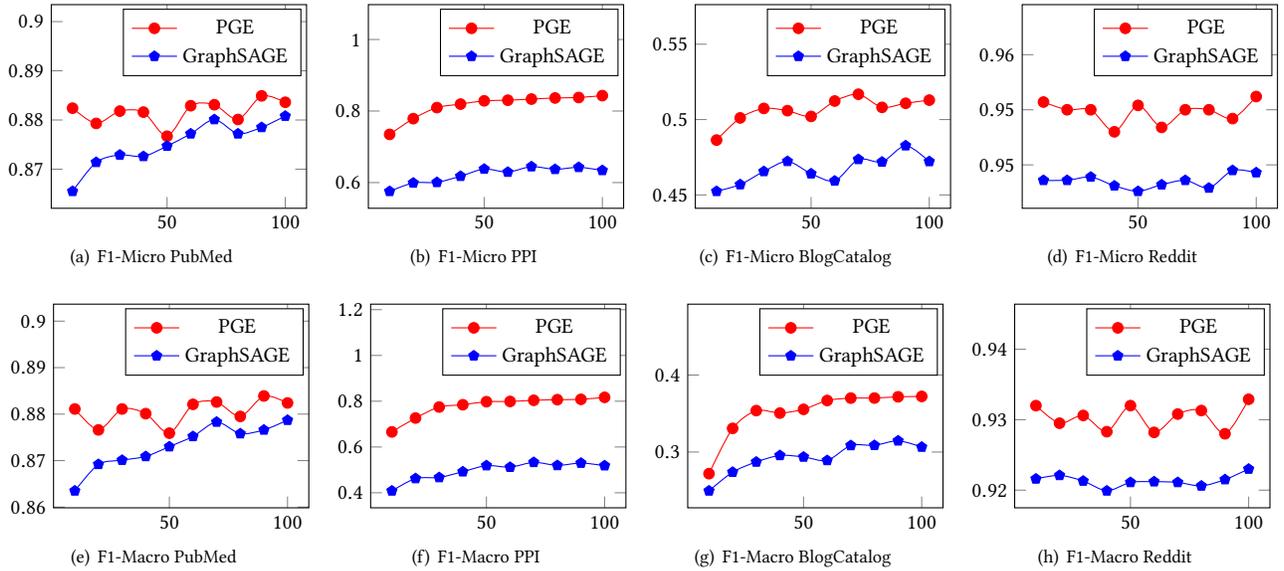
\begin{figure*}[htbp]
	{\small 
	\centering
	\subfigure[F1-Micro PubMed]{
		\pgfplotsset{width=4.975cm}
		\begin{tikzpicture}
		\begin{axis}
		\addlegendentry{PGE}
		\addplot[color=red, mark=*, smooth]
		coordinates
		{(10, 0.8824) (20, 0.8793) (30, 0.8818) (40, 0.8816) (50, 0.8767) (60, 0.8829) (70, 0.8831) (80, 0.8801) (90, 0.8849) (100, 0.8836)};
		\addlegendentry{GraphSAGE}
		\addplot[color=blue,mark=pentagon*, smooth]
		coordinates
		{(10, 0.8655) (20, 0.8714) (30, 0.8729) (40, 0.8726) (50, 0.8747) (60, 0.8772) (70, 0.8801) (80, 0.8772) (90, 0.8785) (100, 0.8808)};
		\addplot[color=blue,mark=pentagon*, smooth]
		coordinates
		{(90, 0.9)};
		\end{axis}
		\end{tikzpicture}
	}
	\subfigure[F1-Micro PPI]{
		\pgfplotsset{width=4.975cm}
		\begin{tikzpicture}
		\begin{axis}
		\addlegendentry{PGE}
		\addplot[color=red, mark=*, smooth]
		coordinates
		{(10, 0.7347) (20, 0.7787) (30, 0.8096) (40, 0.8194) (50, 0.8285) (60, 0.8302) (70, 0.8334) (80, 0.8365) (90, 0.8378) (100, 0.8431)};
		\addlegendentry{GraphSAGE}
		\addplot[color=blue,mark=pentagon*, smooth]
		coordinates
		{(10, 0.5756) (20, 0.5991) (30, 0.6007) (40, 0.6178) (50, 0.6379) (60, 0.6293) (70, 0.6447) (80, 0.6371) (90, 0.6427) (100, 0.6341)};
		\addplot[color=blue,mark=pentagon*, smooth]
		coordinates
		{(90, 1.05)};
		\end{axis}
		\end{tikzpicture}
	}
	\subfigure[F1-Micro BlogCatalog]{
		\pgfplotsset{width=4.975cm}
		\begin{tikzpicture}
		\begin{axis}
		\addlegendentry{PGE}
		\addplot[color=red, mark=*, smooth]
		coordinates
		{(10, 0.4864) (20, 0.5011) (30, 0.5074) (40, 0.5059) (50, 0.5021) (60, 0.5123) (70, 0.5168) (80, 0.5081) (90, 0.5108) (100, 0.5129)};
		\addlegendentry{GraphSAGE}
		\addplot[color=blue,mark=pentagon*, smooth]
		coordinates
		{(10, 0.4524) (20, 0.4569) (30, 0.4656) (40, 0.4723) (50, 0.4641) (60, 0.4593) (70, 0.4736) (80, 0.4719) (90, 0.4827) (100, 0.4722)};
		\addplot[color=blue,mark=pentagon*, smooth]
		coordinates
		{(90, 0.565)};
		\end{axis}
		\end{tikzpicture}
	}
	\subfigure[F1-Micro Reddit]{
		\pgfplotsset{width=4.975cm}
		\begin{tikzpicture}
		\begin{axis}
		\addlegendentry{PGE}
		\addplot[color=red, mark=*, smooth]
		coordinates
		{(10, 0.9557) (20, 0.9550) (30, 0.9550) (40, 0.9530) (50, 0.9554) (60, 0.9534) (70, 0.9550) (80, 0.9550) (90, 0.9542) (100, 0.9562)};
		\addlegendentry{GraphSAGE}
		\addplot[color=blue,mark=pentagon*, smooth]
		coordinates
		{(10, 0.9486) (20, 0.9486) (30, 0.9489) (40, 0.9481) (50, 0.9476) (60, 0.9482) (70, 0.9486) (80, 0.9479) (90, 0.9495) (100, 0.9493)};
		\addplot[color=blue,mark=pentagon*, smooth]
		coordinates
		{(90, 0.963)};
		\end{axis}
		\end{tikzpicture}
	}
	
	\subfigure[F1-Macro PubMed]{
		\pgfplotsset{width=4.975cm}
		\begin{tikzpicture}
		\begin{axis}
		\addlegendentry{PGE}
		\addplot[color=red, mark=*, smooth]
		coordinates
		{(10, 0.8811) (20, 0.8766) (30, 0.8811) (40, 0.8801) (50, 0.8759) (60, 0.8821) (70, 0.8826) (80, 0.8795) (90, 0.8839) (100, 0.8824)};
		\addlegendentry{GraphSAGE}
		\addplot[color=blue,mark=pentagon*, smooth]
		coordinates
		{(10, 0.8635) (20, 0.8692) (30, 0.8701) (40, 0.8709) (50, 0.8730) (60, 0.8752) (70, 0.8783) (80, 0.8758) (90, 0.8766) (100, 0.8787)};
		\addplot[color=blue,mark=pentagon*, smooth]
		coordinates
		{(90, 0.9)};
		\end{axis}
		\end{tikzpicture}
	}
	\subfigure[F1-Macro PPI]{
		\pgfplotsset{width=4.975cm}
		\begin{tikzpicture}
		\begin{axis}
		\addlegendentry{PGE}
		\addplot[color=red, mark=*, smooth]
		coordinates
		{(10, 0.6659) (20, 0.7266) (30, 0.7746) (40, 0.7846) (50, 0.7979) (60, 0.7989) (70, 0.8040) (80, 0.8068) (90, 0.8082) (100, 0.8169)};
		\addlegendentry{GraphSAGE}
		\addplot[color=blue,mark=pentagon*, smooth]
		coordinates
		{(10, 0.4085) (20, 0.4624) (30, 0.4668) (40, 0.4921) (50, 0.5188) (60, 0.5118) (70, 0.5331) (80, 0.5196) (90, 0.5299) (100, 0.5185)};
		\addplot[color=blue,mark=pentagon*, smooth]
		coordinates
		{(90, 1.15)};
		\end{axis}
		\end{tikzpicture}
	}
	\subfigure[F1-Macro BlogCatalog]{
		\pgfplotsset{width=4.975cm}
		\begin{tikzpicture}
		\begin{axis}
		\addlegendentry{PGE}
		\addplot[color=red, mark=*, smooth]
		coordinates
		{(10, 0.2719) (20, 0.3307) (30, 0.3538) (40, 0.3506) (50, 0.3554) (60, 0.3669) (70, 0.3700) (80, 0.3701) (90, 0.3717) (100, 0.3722)};
		\addlegendentry{GraphSAGE}
		\addplot[color=blue,mark=pentagon*, smooth]
		coordinates
		{(10, 0.2496) (20, 0.2739) (30, 0.2871) (40, 0.2956) (50, 0.2936) (60, 0.2891) (70, 0.3085) (80, 0.3087) (90, 0.3146) (100, 0.3065)};
		\addplot[color=blue,mark=pentagon*, smooth]
		coordinates
		{(90, 0.47)};
		\end{axis}
		\end{tikzpicture}
	}
	\subfigure[F1-Macro Reddit]{
		\pgfplotsset{width=4.975cm}
		\begin{tikzpicture}
		\begin{axis}
		\addlegendentry{{\small PGE}}
		\addplot[color=red, mark=*, smooth]
		coordinates
		{(10, 0.9320) (20, 0.9295) (30, 0.9306) (40, 0.9283) (50, 0.9320) (60, 0.9282) (70, 0.9308) (80, 0.9313) (90, 0.9280) (100, 0.9329)};
		\addlegendentry{{\small GraphSAGE}}
		\addplot[color=blue,mark=pentagon*, smooth]
		coordinates
		{(10, 0.9216) (20, 0.9221) (30, 0.9213) (40, 0.9199) (50, 0.9211) (60, 0.9212) (70, 0.9211) (80, 0.9206) (90, 0.9215) (100, 0.9230)};
		\addplot[color=blue,mark=pentagon*, smooth]
		coordinates
		{(90, 0.944)};
		\end{axis}
		\end{tikzpicture}
	}
	\vspace{-2mm}
	\caption{The effects of epoch number}
	\label{f1_pic}
		}
\end{figure*}

\begin{figure*}[ht]
	\centering
	\includegraphics[scale=0.57]{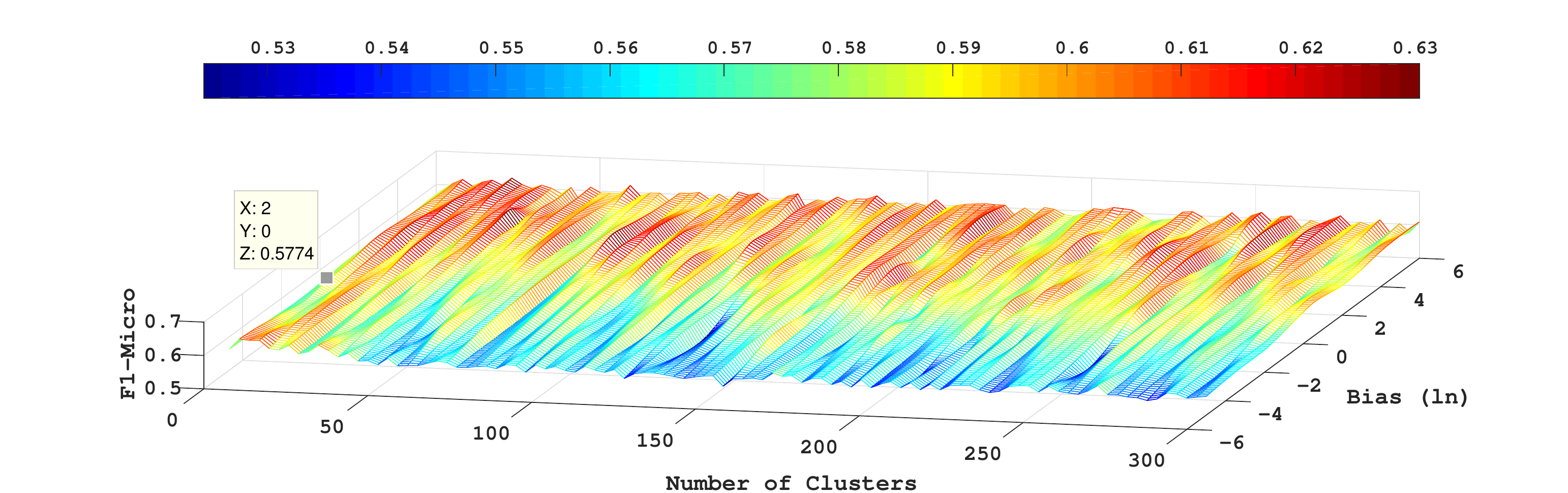}
			\vspace{-3mm}
	\caption{The effects of bias values and cluster number (best viewed as 2D color images)}
	\label{fig:ppi}
\end{figure*}

The Reddit graph is much larger than the other graphs, and thus we used $K$-Means (with $k=40$) for clustering Reddit instead of using DBSCAN which is much slower. We do not report the results for DeepWalk and node2vec as their training processes did not finish in 10 hours while GraphSAGE and PGE finished in several minutes. We also do not report GCN since it needs to load the full graph matrix into each GPU and ran out of memory on our GPUs (each with 12GB memory). PGE's F1-scores are  about 1\% higher than those of GraphSAGE, which we believe is a significant improvement given that the accuracy of GraphSAGE is already very high (94.93\% and 92.30\%).

\subsection{Link Prediction}\label{exp:linkp}

Next we evaluate the quality of graph embedding for link prediction. Given two nodes' embeddings $\textbf{z}_{v}$ and $\textbf{z}_{v'}$, the model should predict whether there is a potential edge existing between them. We used MRR (mean reciprocal rank)~\cite{mrr} to evaluate the performance of link prediction. Specifically, for a node $v$ and $|Q|$ sets of nodes to be predicted, the MRR score can be calculated by the set of prediction queries/lists in $Q$ with $\frac{1}{|Q|} \sum_{i=1}^{|Q|} \frac{1}{rank_i}$, where $rank_i$ is the place of the first correct prediction. We compared PGE with GraphSAGE as we did not find the evaluation method for link prediction in DeepWalk, node2vec and GCN. For the sparse citation graph PubMed, we set the epoch number to $10$ to avoid the data insufficiency problem. For other datasets, the epoch number was set to $1$. As for the biases $b_d$ and $b_s$ and the clustering methods, they are the same as in the node classification experiment in Section~\ref{exp:nodec}.



Table~\ref{tab:mrr} reports the MRR scores of PGE and GraphSAGE for the four datasets. We also created a variant of PGE by only considering bias (i.e., the edge information was not used). The results show that without considering the edge information, PGE records lower MRR scores than GraphSAGE except for PPI. However, when the edge information is incorporated, PGE significantly outperforms GraphSAGE in all cases and the MRR score of PGE is at least 37\% higher than that of GraphSAGE. According to the MRR score definition, the correct prediction made by PGE is $1$ to $3$ positions ahead of that made by GraphSAGE. Compared with the improvements made by PGE for node classification, its improvements for link prediction are much more convincing, which can be explained as follows. Differentiating between neighboring nodes may not have a direct effect on predicting a link between two nodes; rather, the use of edge information by PGE makes a significant difference compared with GraphSAGE and the variant of PGE, as the latter two do not use edge information. 

\subsection{Parameter Sensitivity Tests} \label{exp:params}

In this set of experiments, we evaluated the effects of the parameters in PGE on its performance. 


\subsubsection{Effects of the Epoch Number}


To test the effects of the number of training epochs, we compared PGE with GraphSAGE by varying the epoch number from 10 to 100.  We report the F1-Micro and F1-Macro scores for node classification on the four datasets in Figure~\ref{f1_pic}. The results show that PGE and GraphSAGE have similar trends in F1-Micro and F1-Marco, although PGE always outperforms GraphSAGE.  Note that the training time increases linearly with the epoch number, but the training time for 100 epochs is also only tens of seconds (for the small dataset PubMed) to less than 5 minutes (for the largest dataset Reddit).

\subsubsection{Effects of Biases and Cluster Number} \label{exp:bias}

We also tested the effects of different bias values and the number of clusters. We ran PGE for 1,000 times for node classification on PPI, using different number of clusters $k$ and different values of $b_d$ (by fixing $b_s=1$). We used $K$-Means for Step 1 since it is flexible to change the value $k$. The number of training epochs was set at 10 for each run of PGE. All the other parameters were set as their default values. 


Figure~\ref{fig:ppi} reports the results, where the $X$-axis shows the number of clusters $k$, the $Y$-axis indicates the logarithmic value (with the base $e$) of $b_d$, and the $Z$-axis is the F1-Micro score (F1-Macro score is similar and omitted). The results show that taking a larger bias $b_d$ (i.e., $Y > 0$) can bring positive influence on the F1-score independent of the cluster number $k$, and the performance increases as a larger $b_d$ is used. When $b_d$ is less than $1$,  i.e., $b_d < b_s$, it does not improve the performance over uniform neighbor sampling (i.e., $b_d = b_s$ or $Y=0$). This indicates that selecting a larger number of dissimilar neighbors (as a larger $b_d$ means a higher probability of including dissimilar neighbors into $\mathcal{G}^s$) helps improve the quality of node embedding, which is consistent with our analysis in Section~\ref{sec:analysis}.

For the number of clusters $k$, as the average degree of the PPI graph is $14.38$, when the cluster number is more than $50$,  the F1-score becomes fluctuating  to $k$ (i.e., the shape is like waves in Figure~\ref{fig:ppi}). This phenomenon is caused by the limitation of the clustering algorithm, since $K$-Means is sensitive to noises and a large $k$ is more likely to be affected by noises. Note that when the cluster number is not large (less than $50$), a small bias $b_d$ (less than $1$) may also improve the F1-score, which may be explained by the fact that there are homophily and structural equivalence features in the graph, while $b_d < 1$ indicates that nodes tend to select similar neighbors to aggregate. In general, however, a large $b_d$ and a small cluster number $k$ (close to the average degree) are more likely to improve the performance of the neighborhood aggregation method.

\section{Conclusions}	\label{sec:conclusions}



We presented a representation learning framework, called PGE, for property graph embedding. The key idea of PGE is a three-step procedure to leverage both the topology and property information to obtain a better node embedding result. Our experimental results validated that, by incorporating the richer information contained in a property graph into the embedding procedure, PGE achieves better performance than existing graph embedding methods such as DeepWalk~\cite{DeepWalk}, node2vec~\cite{node2vec}, GCN~\cite{features_nn_3} and GraphSAGE~\cite{graphsage}. PGE is a key component in the GNN library of MindSpore --- a unified training and inference framework for device, edge, and cloud in Huawei’s full-stack, all-scenario AI portfolio --- and has a broad range of applications such as recommendation in Huawei’s mobile services, cloud services and 5G IoT applications.   

\vspace{2mm}

\noindent \textbf{Acknowledgments.} We thank the reviewers for their valuable comments. We also thank Dr. Yu Fan from Huawei for his contributions on the integration of PGE into MindSpore and its applications.  This work was supported in part by ITF 6904945 and GRF 14222816.

\bibliography{graphgen}

\begin{thebibliography}{10}

\bibitem{blogcatalog}
Nitin Agarwal, Huan Liu, Sudheendra Murthy, Arunabha Sen, and Xufei Wang.
\newblock A social identity approach to identify familiar strangers in a social
  network.
\newblock In {\em ICWSM}, 2009.

\bibitem{factorization_GF}
Amr Ahmed, Nino Shervashidze, Shravan~M. Narayanamurthy, Vanja Josifovski, and
  Alexander~J. Smola.
\newblock Distributed large-scale natural graph factorization.
\newblock In {\em WWW}, pages 37--48, 2013.

\bibitem{homophily_reason_2}
Eytan Bakshy, Itamar Rosenn, Cameron Marlow, and Lada~A. Adamic.
\newblock The role of social networks in information diffusion.
\newblock In {\em WWW}, pages 519--528, 2012.

\bibitem{homophily_reason_1}
Mauro Barone and Michele Coscia.
\newblock Birds of {A} feather scam together: Trustworthiness homophily in {A}
  business network.
\newblock {\em Social Networks}, 54.

\bibitem{representation_learning}
Yoshua Bengio, Aaron~C. Courville, and Pascal Vincent.
\newblock Representation learning: {A} review and new perspectives.
\newblock {\em PAMI}, 35:1798--1828, 2013.

\bibitem{introduction_degree}
Smriti Bhagat, Graham Cormode, and S.~Muthukrishnan.
\newblock Node classification in social networks.
\newblock In {\em Social network datanalytics}, pages 115--148. 2011.

\bibitem{euclidean_data}
Michael~M. Bronstein, Joan Bruna, Yann LeCun, Arthur Szlam, and Pierre
  Vandergheynst.
\newblock Geometric deep learning: Going beyond euclidean data.
\newblock {\em IEEE Signal Processing Magazine}, 34.

\bibitem{factorization_GraRep}
Shaosheng Cao, Wei Lu, and Qiongkai Xu.
\newblock Grarep: Learning graph representations with global structural
  information.
\newblock In {\em CIKM}, pages 891--900, 2015.

\bibitem{DNGR}
Shaosheng Cao, Wei Lu, and Qiongkai Xu.
\newblock Deep neural networks for learning graph representations.
\newblock In {\em AAAI}, pages 1145--1152, 2016.

\bibitem{HARP}
Haochen Chen, Bryan Perozzi, Yifan Hu, and Steven Skiena.
\newblock {HARP:} hierarchical representation learning for networks.
\newblock In {\em AAAI}, pages 2127--2134, 2018.

\bibitem{NB-GCN}
Jianfei Chen, Jun Zhu, and Le~Song.
\newblock Stochastic training of graph convolutional networks with variance
  reduction.
\newblock In {\em ICML}, pages 941--949, 2018.

\bibitem{DBSCAN}
Martin Ester, Hans{-}Peter Kriegel, J{\"{o}}rg Sander, and Xiaowei Xu.
\newblock A density-based algorithm for discovering clusters in large spatial
  databases with noise.
\newblock In {\em SIGKDD}, pages 226--231, 1996.

\bibitem{structural_equivalent_reason_1}
Santo Fortunato.
\newblock Community detection in graphs.
\newblock {\em Physics Reports}, 486:75--174, 2010.

\bibitem{introduction_kernel}
Thomas G{\"{a}}rtner, Tam{\'{a}}s Horv{\'{a}}th, and Stefan Wrobel.
\newblock Graph kernels.
\newblock In {\em Encyclopedia of Machine Learning and Data Mining}, pages
  579--581. 2017.

\bibitem{graph_embedding}
Palash Goyal and Emilio Ferrara.
\newblock Graph embedding techniques, applications, and performance: {A}
  survey.
\newblock {\em KBS}, 151:78--94, 2018.

\bibitem{node2vec}
Aditya Grover and Jure Leskovec.
\newblock node2vec: Scalable feature learning for networks.
\newblock In {\em SIGKDD}, pages 855--864, 2016.

\bibitem{overview_Standford}
William~L. Hamilton, Rex Ying, and Jure Leskovec.
\newblock Representation learning on graphs: Methods and applications.
\newblock {\em {IEEE} Data Eng. Bull.}, 40:52--74, 2017.

\bibitem{graphsage}
William~L. Hamilton, Zhitao Ying, and Jure Leskovec.
\newblock Inductive representation learning on large graphs.
\newblock In {\em NIPS}, pages 1024--1034, 2017.

\bibitem{structural_equivalent_reason_2}
Keith Henderson, Brian Gallagher, Tina Eliassi{-}Rad, Hanghang Tong, Sugato
  Basu, Leman Akoglu, Danai Koutra, Christos Faloutsos, and Lei Li.
\newblock Rolx: Structural role extraction {\&} mining in large graphs.
\newblock In {\em SIGKDD}, pages 1231--1239, 2012.

\bibitem{deep_neural_network}
Geoffrey~E. Hinton and Ruslan~R. Salakhutdinov.
\newblock Reducing the dimensionality of data with neural networks.
\newblock {\em Science}, 313:504--507, 2006.

\bibitem{adam}
Diederik~P. Kingma and Jimmy Ba.
\newblock Adam: {A} method for stochastic optimization.
\newblock In {\em ICLR}, 2015.

\bibitem{features_nn_3}
Thomas~N. Kipf and Max Welling.
\newblock Semi-supervised classification with graph convolutional networks.
\newblock {\em CoRR}, 2016.

\bibitem{features_nn_1}
Thomas~N. Kipf and Max Welling.
\newblock Variational graph auto-encoders.
\newblock {\em CoRR}, 2016.

\bibitem{introduction_local_neighborhood}
David Liben{-}Nowell and Jon~M. Kleinberg.
\newblock The link-prediction problem for social networks.
\newblock {\em {JASIST}}, 58:1019--1031, 2007.

\bibitem{k_means}
James MacQueen.
\newblock Some methods for classification and analysis of multivariate
  observations.
\newblock In {\em Berkeley Symposium on Mathematical Statistics and
  Probability}, volume~1, pages 281--297, 1967.

\bibitem{skip_gram}
Tomas Mikolov, Ilya Sutskever, Kai Chen, Gregory~S. Corrado, and Jeffrey Dean.
\newblock Distributed representations of words and phrases and their
  compositionality.
\newblock In {\em NIPS}, pages 3111--3119, 2013.

\bibitem{pubmed}
Galileo Namata, Ben London, Lise Getoor, Bert Huang, and UMD EDU.
\newblock Query-driven active surveying for collective classification.
\newblock In {\em MLG}, 2012.

\bibitem{relational_machine_learning}
Maximilian Nickel, Kevin Murphy, Volker Tresp, and Evgeniy Gabrilovich.
\newblock A review of relational machine learning for knowledge graphs.
\newblock {\em Proceedings of the {IEEE}}, 104.

\bibitem{factorization_HOPE}
Mingdong Ou, Peng Cui, Jian Pei, Ziwei Zhang, and Wenwu Zhu.
\newblock Asymmetric transitivity preserving graph embedding.
\newblock In {\em SIGKDD}, pages 1105--1114, 2016.

\bibitem{DeepWalk}
Bryan Perozzi, Rami Al{-}Rfou, and Steven Skiena.
\newblock Deepwalk: Online learning of social representations.
\newblock In {\em SIGKDD}, pages 701--710, 2014.

\bibitem{Walklets}
Bryan Perozzi, Vivek Kulkarni, and Steven Skiena.
\newblock Walklets: Multiscale graph embeddings for interpretable network
  classification.
\newblock {\em CoRR}, 2016.

\bibitem{mrr}
Dragomir~R. Radev, Hong Qi, Harris Wu, and Weiguo Fan.
\newblock Evaluating web-based question answering systems.
\newblock In {\em LREC}, 2002.

\bibitem{f1_score}
Yutaka Sasaki.
\newblock The truth of the f-measure.
\newblock {\em Teach Tutor Mater}, 1:1--5, 2007.

\bibitem{features_nn_2}
Michael~Sejr Schlichtkrull, Thomas~N. Kipf, Peter Bloem, Rianne van~den Berg,
  Ivan Titov, and Max Welling.
\newblock Modeling relational data with graph convolutional networks.
\newblock In {\em ESWC}, pages 593--607, 2018.

\bibitem{ppi}
Chris Stark, Bobby{-}Joe Breitkreutz, Teresa Reguly, Lorrie Boucher, Ashton
  Breitkreutz, and Mike Tyers.
\newblock Biogrid: {A} general repository for interaction datasets.
\newblock {\em Nucleic Acids Research}, 34:535--539, 2006.

\bibitem{SDNE}
Daixin Wang, Peng Cui, and Wenwu Zhu.
\newblock Structural deep network embedding.
\newblock In {\em SIGKDD}, pages 1225--1234, 2016.

\bibitem{structural_equivalent_reason_3}
Jaewon Yang and Jure Leskovec.
\newblock Overlapping communities explain core-periphery organization of
  networks.
\newblock {\em Proceedings of the {IEEE}}, 102.

\end{thebibliography}
\end{document}